\documentclass{article} 
\usepackage{nips13submit_e,times}
\usepackage{hyperref}
\usepackage{url}
\usepackage{amssymb}
\usepackage{amsmath}
\usepackage{epsfig}
\usepackage{float}
\usepackage{multirow}
\usepackage{graphicx}
\usepackage{setspace}
\usepackage{subfigure}
\usepackage{umoline}

\title{Pay More Attention: Neural Architectures for Question-Answering}

\author{
Zia Hasan\\
Samsung Electronics\\
Suwon, South Korea\\
\texttt{ziahasan@stanford.edu} \\
\And
Sebastian Fischer\\
Deutsche Telekom AG \\
Berlin, Germany \\
\texttt{seb1988@stanford.edu} \\
}

\nipsfinalcopy 

\begin{document}

\maketitle

\begin{abstract}
Machine comprehension is a representative task of natural language understanding. Typically, we are given context paragraph and the objective is to answer a question that depends on the context. Such a problem requires to model the complex interactions between the context paragraph and the question. Lately, attention mechanisms have been found to be quite successful at these tasks and in particular, attention mechanisms with attention flow from both context-to-question and question-to-context have been proven to be quite useful. In this paper, we study two state-of-the-art attention mechanisms called Bi-Directional Attention Flow (BiDAF) and Dynamic Co-Attention Network (DCN) and propose a hybrid scheme combining these two architectures that gives better overall performance. Moreover, we also suggest a new simpler attention mechanism that we call Double Cross Attention (DCA) that provides better results compared to both BiDAF and Co-Attention mechanisms while providing similar performance as the hybrid scheme. The objective of our paper is to focus particularly on the attention layer and to suggest improvements on that. Our experimental evaluations show that both our proposed models achieve superior results on the Stanford Question Answering Dataset (SQuAD) compared to BiDAF and DCN attention mechanisms.
\end{abstract}

\section{Introduction}
Enabling machines to understand natural language is one of the key challenges to achieve artificially intelligent systems. Asking machines questions and getting a meaningful answer adds value to us since it automatizes knowledge acquisition efforts drastically. Apple's Siri and Amazon's Echo are two such examples of mass market products capable of machine comprehension that has led to a paradigm shift on how consumers' interact with machines.
 
Over the last decade, research in the field of Natural Language Processing (NLP) has massively benefited from neural architectures. Those approaches have outperformed former state-of-the-art non-neural machine learning model families while needing far less human intervention since they don't require any manual feature engineering. A subset of NLP research focuses on building systems that are able to answer questions about a given document. To jointly expand the current best practice, the Stanford Question Answering Dataset (SQuAD) was setup as a basis for a global competition between different research groups \cite{squad}. SQuAD was published in 2016 and includes 100,000+ context-question-triplets on 500+ articles, significantly larger than previous reading comprehension datasets \cite{squad2}. The context paragraphs were obtained from more then 500 Wikipedia articles and the answers were sourced with Amazon Mechanical Turk. Recently, researchers were able to make machines outperform humans (as of Jan 2018) \cite{squad2}. Answers in this dataset are taken from the document itself and are not dynamically generated from scratch. Instead of generating text that provides a suitable answer, the objective is to find the boundaries in which the answer is contained in the document. The aim is to achieve close to human performance in generating correct answers from a context paragraph given any new unseen questions.

To solve this problem of question answering, neural attention mechanisms have recently gained significant popularity by focusing on the most relevant area within a context paragraph, useful to answer the question \cite{weston, xiong}. Attention mechanisms have proven to be an important extension to achieve better results in NLP problems \cite{mitry}.
While earlier attention mechanisms for this task were usually uni-directional, obtaining a fixed size vector for each context word summarizing the question words, bi-directional attention flow applies an attention scheme in both directions (context-to-question as well as question-to-context). In this paper, we study two state-of-the-art neural architectures with an attention flow going in both directions called Bi-Directional Attention Flow (BiDAF) \cite{bidaf} and Dynamic Co-Attention network (DCN) \cite{coatten} that were once themselves leading architectures in the SQuAD challenge. We would also like to propose yet another hybrid neural architecture that shows competitive results by bringing together these two models. More specifically, we combined the attention layer of both BiDAF and Co-Attention models. In addition to this, we propose another simpler model family called Double Cross Attention (DCA) which in itself performs better than both BiDAF and Co-Attention while giving similar performance as hybrid model. {\bf The objective of this paper is to do a comparative study of the performance of attention layer and not to optimize the performance of the overall system}. 
 
\section{Model}
We started our development by re-implementing the BiDAF and DCN models. We figured that these models individually enhanced the baseline performance significantly, so the hope was that a combination would eventually lead to superior results. Thereby we created our ``hybrid" model, which we will subsequently explain shortly. In the following subsections, we describe each layer of our model in more detail.

\subsection{Word and Character Embedding Layer}
The word embedding layer maps each word in a context and question to a vector of fixed size using pre-trained GloVe embeddings \cite{glove}. First, we encode each word in the question and context with the pre-trained Glove embedding as given in the baseline code. Then we concatenate to the word encodings an optional {\em Character-level Embedding} with CNNs since it helps to deal with out-of-vocabulary words \cite{bidaf, kim}. The joint concatenated encoding of words and characters is subsequently fed into the context and question encoding layer.

\subsection{Context and Question Encoding Layer}
Once we have a context and question embeddings, we use a Bidirectional GRU to translate these context and question embeddings into encodings. Whereas a simple LSTM/GRU cell encodes sequence data such as a sentences only from left-to-right, a bi-directional approach also parses a sentence from the end to the start. Both representations of a sequence are then usually concatenated and are assumed to encode the sequence structure more expressively ultimately leading to higher model performance.

\subsection{Attention Layer}
The attention layer is the modeling layer that eventually involves modeling the complex interactions between the context and question words. Next, we describe several different attention mechanisms that we implemented in our system. 

\subsubsection{Bidirectional attention flow}
We implemented a complete BiDAF layer as suggested in the project handout and in the original paper \cite{bidaf}. Bi-directional attention flow approaches the machine comprehension challenge slightly differently. Instead of using an attention layer for transforming context inputs to fixed-size vectors, the BiDAF model computes the attention from both question-to-context as well as context-to-question and combines them effectively.
The basic idea is essentially to obtain a similarity matrix to capture relations between context and question words and use this matrix to obtain context-to-question as well as question-to-context attention vectors. Finally, these attention vectors are concatenated to the context encodings in a specific way to obtain the output of the Bi-directional attention flow layer. In the original BiDAF paper, an additional Bidirectional-RNN is used to again encode these concatenated vectors. However, it didn't give any improvement in our setup, hence we chose to omit it in our final implementation.

\subsubsection{Dynamic Co-Attention}
Dynamic Co-Attention Network layer (DCN), similar to BiDAF involves a two-way attention between the context and the question but unlike BiDAF, DCN involves a second-level attention computation over the previously computed attentions \cite{coatten}. The dynamic co-attention network (DCN) is an end-to-end neural network architecture. The authors claim that the ability of attending to context inputs strongly depends on the query (question). The intuition behind that is also reflected by a human's ability to better answer a question on an input paragraph, when the question is known before reading the context itself, because then one can attend specifically to relevant information in the context. For details, please check the project handout, the original paper and our implementation code. In the original paper and the project handout, there was also a concept of sentinel vectors that was introduced but in our tests, it again didn't seem to provide any significant advantage, so we again chose to omit this as well in our final implementation. 

\subsubsection{Hybrid BiDAF-Co-Attention (New Model)}
This is model that we propose and it builds heavily on aspects of the BiDAF\cite{bidaf} as well as the DCN models\cite{coatten}.  Since the attention outputs from both the BiDAF and DCN seem to have their merits, our idea was to combine them by concatenating both attentions to the original context states. The intuition was that the neural network should be able to train in order to use and pick them both effectively. Experimental results that we describe later, also verify our claim. Please check the code for exact implementation details

\subsubsection{Double Cross Attention (New Model)}
\begin{figure}
 \includegraphics[width=6.0in]{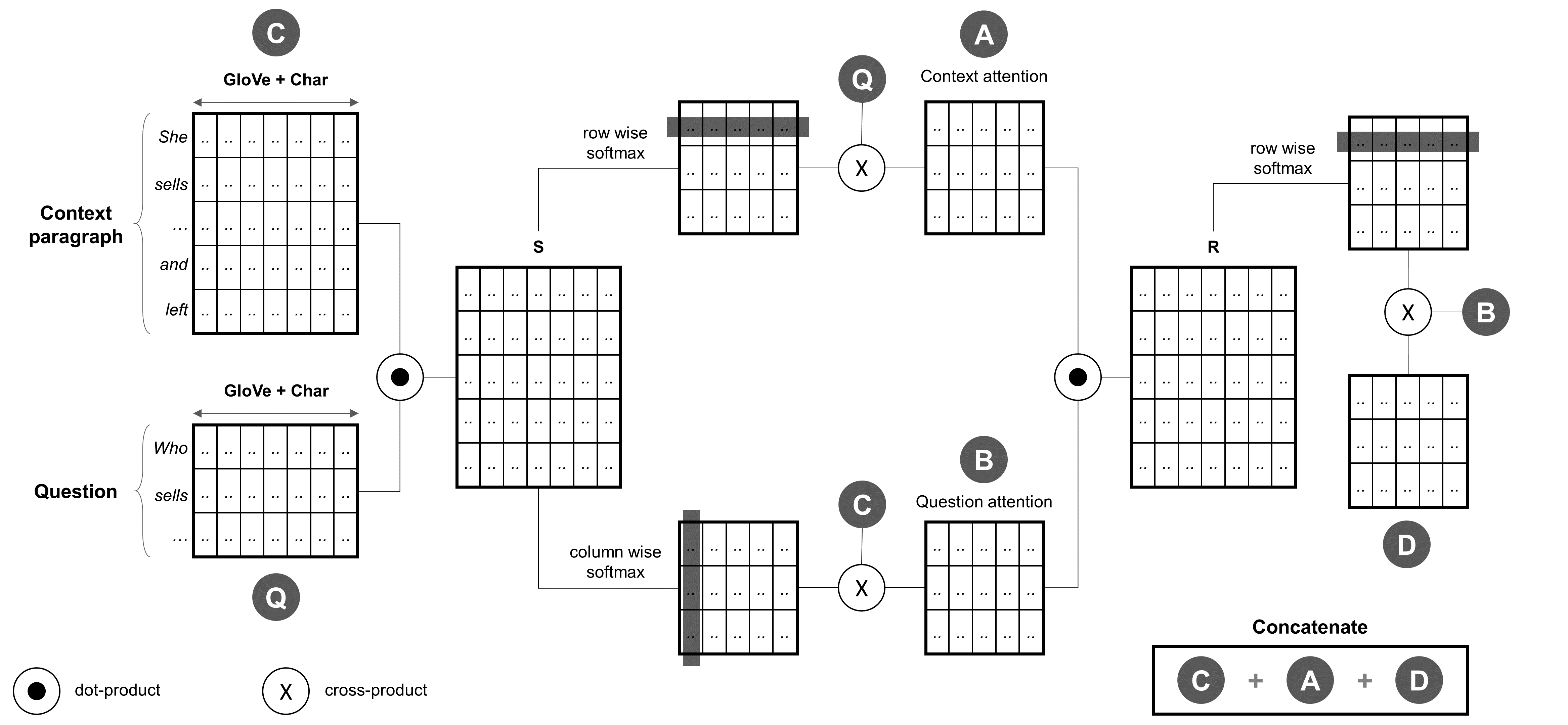}
  \caption{Double Cross Attention Model}
  \label{fig:dca}
\end{figure}

In this section, we propose another simple idea called Double Cross Attention (DCA) which seem to provide better results compared to both BiDAF and Co-Attention while providing similar performance as concatenated hybrid model discussed in previous section. The motivation behind this approach is that first we pay attention to each context and question and then we attend those attentions with respect to each other in a slightly similar way as DCN. The intuition is that if iteratively read/attend both context and question, it should help us to search for answers easily. The DCA mechanism is explained graphically in Figure. \ref{fig:dca} and the formal description of the layer is as follows.

Assume we have context hidden states $\mathbf{c}_1, \mathbf{c}_2...,\mathbf{c}_N\in \mathbb{R}^{2h}$ and question hidden states $\mathbf{q}_1, \mathbf{q}_2...,\mathbf{q}_M\in \mathbb{R}^{2h}$ obtained after passing context and question embeddings through a bi-directional GRU.  First, we compute a cross-attention matrix  $\mathbf{S}\in\mathbb{R}^{N\times M}$, which contains a similarity score $S_{ij}$ for each pair of context and question hidden states $(\mathbf{c}_i,\mathbf{q}_j)$. We chose  $S_{ij}=\mathbf{c}_i^T\mathbf{q}_j$, since it is a parameter free approach to calculate attention but one can also construct this function with a trainable weight parameter (which can be shared in the subsequent step). 

First we obtain Context-to-Question (C2Q) attention vectors $\mathbf{a}_i$ as follows:
\begin{align}
\alpha_i = \text{softmax} \mathbf{S_{(i:)}}\in \mathbb{R}^M,
\mathbf{a}_i = \sum_{j=1}^{M}\alpha_i^j\mathbf{q}_j \in \mathbb{R}^{2h}
\end{align}

Next, we also obtain Question-to-Context (Q2C) attention vectors $\mathbf{b}_j$ as follows:
\begin{align}
\beta_j = \text{softmax} \mathbf{S_{(:j)}}\in \mathbb{R}^N,
\mathbf{b}_j = \sum_{i=1}^{N}\beta_j^i\mathbf{c}_i \in \mathbb{R}^{2h}
\end{align}

Then we compute a second-level cross attention matrix  $\mathbf{R}\in\mathbb{R}^{N\times M}$, which contains a similarity score $R_{ij}$ for each pair of context and question attention states $(\mathbf{a}_i, \mathbf{b}_j)$. We again choose a simple dot product attention $R_{ij}=\mathbf{a}_i^T\mathbf{b}_j$. Additionally, we obtain Context Attention-to-Question Attention(CA2QA) cross attention vectors $\mathbf{d}_i$ as follows:
\begin{align}
\gamma_i = \text{softmax} \mathbf{R_{(i:)}}\in \mathbb{R}^M,
\mathbf{d}_i = \sum_{1}^{M}\gamma_i^j\mathbf{b}_j \in \mathbb{R}^{2h}
\end{align}

Finally, we concatenate $\mathbf{c}_i$, $\mathbf{a}_i$ and $\mathbf{d}_i$ as a new state $[\mathbf{c}_i; \mathbf{a}_i; \mathbf{d}_i]$ and pass it through a biLSTM layer to obtain double query attended encoded context states as follows.
 \begin{align}
\{\mathbf{u}_1,.... \mathbf{u}_N\} = \text{biLSTM} (\{[\mathbf{c}_1; \mathbf{a}_1; \mathbf{d}_1],.... [\mathbf{c}_N; \mathbf{a}_N; \mathbf{d}_N]\})
\end{align}

Finally all attention layer outputs are concatenated and fed into a Softmax layer that computes the probability distributions for the start and end token independently, as it is done in the baseline implementation.


\section{Experiments}
\begin{figure}[htp]
	\centering
	\includegraphics[width=.5\textwidth]{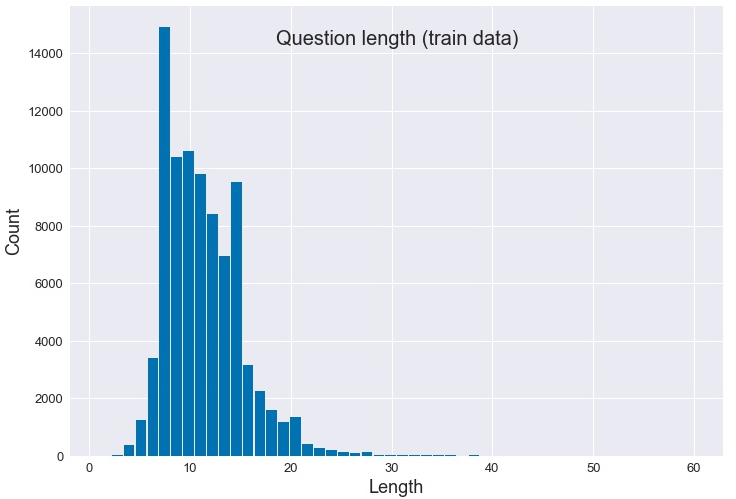}\hfill
	\includegraphics[width=.5\textwidth]{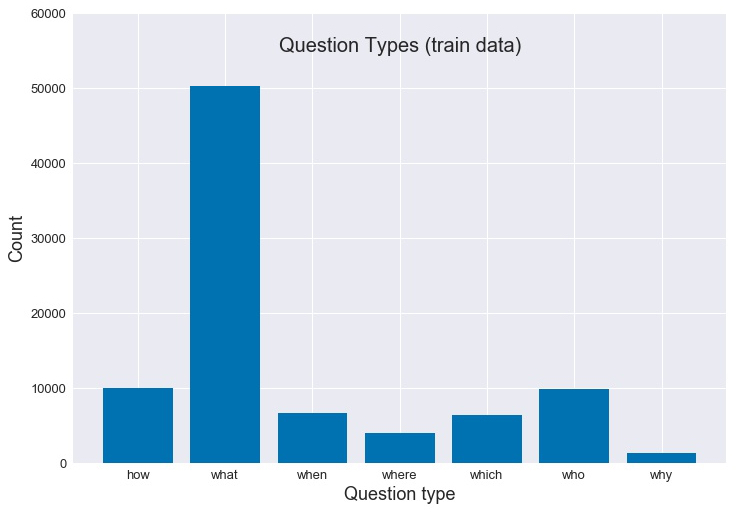}\hfill
	\includegraphics[width=.5\textwidth]{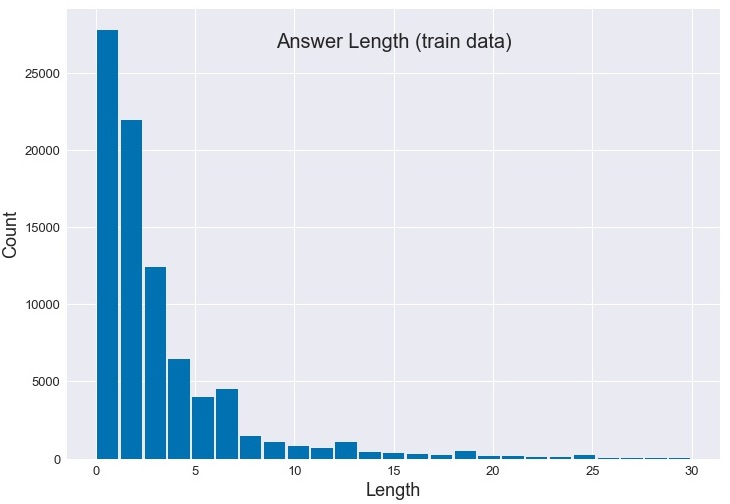}\hfill
	\includegraphics[width=.5\textwidth]{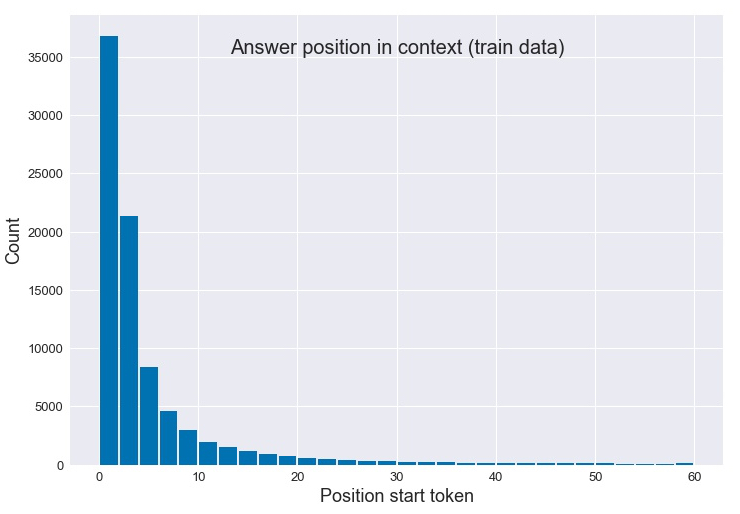}\hfill
	\caption{Exploratory Data Analysis}
	\label{dataset}
\end{figure}
Before we started the enhancements of the baseline model, we studied the SQuAD data set. Figure. \ref{dataset} shows the distribution of the answer, question and context lengths as well as the relative position of the answer span inside a context. Furthermore, we counted the different question types. We found that most answers have a length less than 5 words. Additionally, a question usually consists of 5-20 words. Moreover, we noticed that on average a context is of length ~120 (visualization excluded due to lack of space). Furthermore, answers for a question tend to be represented by spans of context words that are at the beginning of a context. Finally, we can see that ``what" questions build the majority of questions, almost the same amount as all other question types combined.

%

\subsection{Results}
In this section, we report the results of our experiments. To ensure the generality of our model, we used Dropout technique for regularizing neural networks. We start our experiments with default hyperparameters: embedding size of 100, batch size 100, hidden size 200, learning rate of 0.001 and a dropout rate of 0.15. For character level encoding, default character embedding size is 20, kernel size is 5 and number of filters are 100. For each architecture, we report the evaluation metrics F1 and EM (Exact match) computed on the dev set. 

\begin{figure}[htp]
\centering
\includegraphics[width=.5\textwidth]{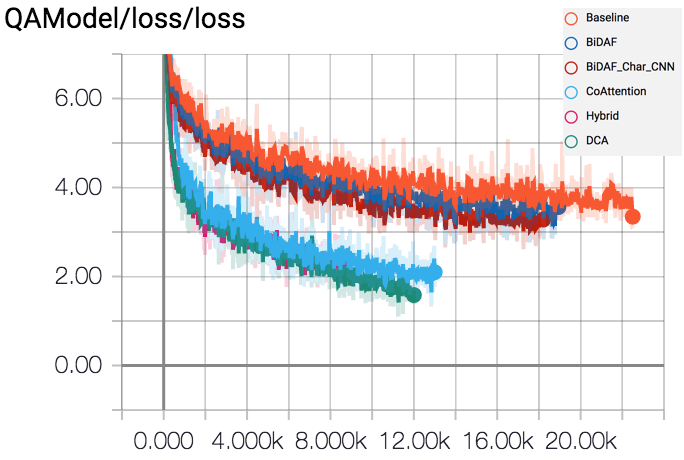}\hfill
\includegraphics[width=.5\textwidth]{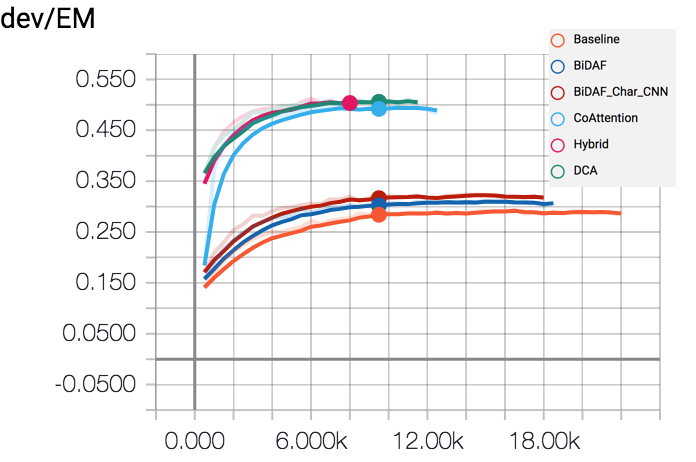}\hfill
\includegraphics[width=.5\textwidth]{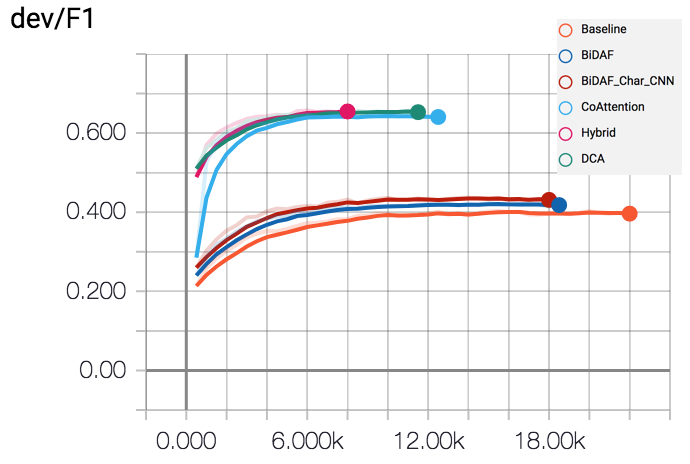}\hfill
\includegraphics[width=.5\textwidth]{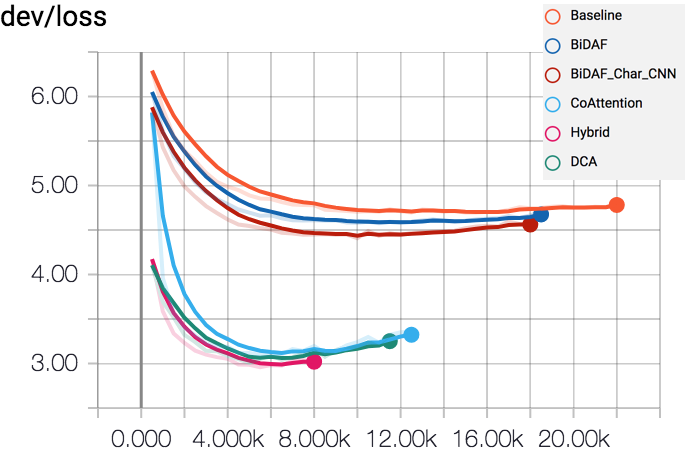}\hfill
\caption{Tensorboard Visualizations}
\label{tensorboard}
\end{figure}

The effect of character embedding on the BiDAF model is reported in Table \ref{tb:charemb}. We can notice that character embedding boosts up the performance by roughly 2\% for both EM and F1 score. This is expected since character embedding can help deal with non-dictionary words by giving them a unique embedding.  Next, we report the results of the model performances for baseline, BiDAF, Co-Attention, Hybrid and DCA attention mechanisms in Table \ref{tb:perresults}. Notice that none of these architectures were optimized for EM/F1 scores but we are more interested in difference between these mechanisms for a fixed set of hyperparameters. Hybrid and DCA have a slight edge over plain BiDAF and Co-Attention module as per the results. Co-Attention with char embedding was giving us worse results so we put the best numbers we got for Co-Attention there. We would like to point out that the BiDAF model here doesn't include BiLSTM layer as present in original paper because the BiLSTM didn't give any advantage except for slowing down the training. Selected tensorboard visualizations are also shown in Figure \ref{tensorboard}. Visualizations demonstrate that both hybrid and DCA models perform better than vanilla Co-Attention and BiDAF attention mechanisms and reduce the losses faster and increase the dev F1/EM scores faster as well.

\begin{table}
\parbox{.45\linewidth}{
\centering
\small
       \begin{tabular}{ p{4cm}  p{1.0cm} p{1.0cm}}
    \bfseries MODEL (BiDAF)   &     \bfseries F1  &     \bfseries  EM \\ \hline
    Char Embedding Disabled & 45.94\% & 36.91\%\\ 
    Char Embedding Enable     & 47.48\% & 38.62\%\\\hline
    \end{tabular}
    \caption{Effect of Character Embedding}
    \label{tb:charemb}
}
\hfill
\parbox{.45\linewidth}{
\centering
\small
       \begin{tabular}{ p{2cm}  p{1.0cm} p{1.0cm}}
    \bfseries MODEL  &     \bfseries F1  &     \bfseries  EM \\ \hline
    Baseline* 			  & 43.44\% & 34.37\%\\ 
    BiDAF    		 	  & 47.48\% & 38.62\%\\
    Co-Attention*    	  	  & 69.56\% & 58.52\%\\
    Hybrid   			  & 70.95\% & 60.54\%\\
    DCA  & 70.68\% & 60.37\%\\
    \hline
    \end{tabular}\\
    \tiny{*no char embedding}   
 \caption{Performance Results}
    \label{tb:perresults}
}
\end{table}
\begin{center}
\begin{table}[ht]
\caption{F1/EM\% on different question types}
{\small
\hfill{}
       \begin{tabular}{ p{1.4cm}  p{1.3cm} p{1.3cm} p{1.3cm} p{1.3cm} p{1.3cm} p{1.3cm} p{1.3cm}}
    \bfseries MODEL  &    \bfseries what  &     \bfseries  how &     \bfseries  who &     \bfseries  when  &     \bfseries  which &     \bfseries  where & \bfseries  why \\ \hline
    Baseline 			  &44.23/34.81 &54.14/42.38 &34.94/31.51 &39.28/29.63 &45.76/38.57 &34.50/26.47 & 48.65/14.29\\ 
    BiDAF    		 	  &50.12/41.51 &57.84/46.36 &42.64/36.99 &44.64/36.42 &51.37/44.71 &39.91/30.88 & 72.06/57.14 \\ 
    DCN    	  	  &71.44/61.53 & 73.83/60.93& 66.24/60.73&66.08/53.08& 76.35/67.57& 68.64/58.82& 82.12/57.14\\ 
    Hybrid   			  &74.83/64.92& 77.71/66.89& 66.29/58.45& 65.30/53.09& 76.52/67.58& 73.22/63.24& 74.05/57.14 \\ 
    DCA   &73.16/63.62& 77.59/65.56& 68.77/61.19& 65.39/51.85& 74.35/64.51& 71.33/61.76& 70.24/57.14 \\ 
    \hline
    \end{tabular}
}
\hfill{}
\label{tb:questiontypes}
\end{table}
\end{center}


%

\subsection{Hyperparameter Tuning}
We made a brief attempt to do a bit of hyperparameter tuning on our proposed DCA model and we report the results in Table \ref{tb:hyper}. Ideally, hyperparameter tuning for neural network architectures should be done using bayesian hyperparameter optimization but due to the lack of time we tried to do a random search on a small set of hyperparameters that we guessed could be more suitable. While, we didn't find any significantly good set of parameters, we noticed that reducing the hidden size has a minor effect on improving the performance. This is probably because it reduces the system complexity which makes the model easier to train.
\begin{center}
\begin{table}[ht]
\caption{Hyperparameter Tuning for DCA Model}
{\small
\hfill{}
       \begin{tabular}{ p{1.5cm}  p{1.8cm} p{1.5cm} p{1.5cm} p{1.5cm} p{1.0cm} p{1.0cm}}
    \bfseries dropout  & \bfseries learning rate  & \bfseries hidden size  &  \bfseries glove  & \bfseries batch-size  &   \bfseries F1  &     \bfseries  EM \\ \hline
    0.15 & 0.001 &200 &  100 &100 			  & 70.68\% & 60.37\%\\ 
    0.15 & 0.01 &100 &  100 &100 			  & 71.01\% & 60.66\%\\ 
    0.25& 0.01&100&100&100 	  		  & 70.88\% & 60.67\%\\
    0.15& 0.001&100&100&100 	  	  & 71.30\% & 60.78\%\\
    0.10 & 0.0001&300&300&50  &66.26\% & 54.83\%\\ 

    \hline
    \end{tabular}
}
\hfill{}
\label{tb:hyper}
\end{table}
\end{center}

\subsection{Error Analysis}
In Table \ref{tb:error}, we briefly provide error analysis on a small sample of results for hybrid and DCA models and try to explain the model behavior.
\begin{center}
\begin{table}[ht]
\caption{Error Analysis on Hybrid and DCA models on some sample examples}
{\small
\hfill{}
     \begin{tabular}{ p{4cm}  p{10cm} }
    \hline
    \bfseries Context 			  & southern california is home to many major business districts. central business districts (cbd) include downtown los angeles, downtown san diego, downtown san bernardino, downtown bakersfield, south coast metro and downtown riverside.\\ 
    \bfseries Question    		  & what is the only district in the cbd to not have "downtown" in it's name?\\
    \bfseries True Answer   	  	  & south coast metro\\
    \bfseries Predicted Answer   	  &  ""(Hybrid model) / central business districts (DCA model)\\
    \bfseries Explanation of Model Behavior   & Hybrid model didn't output any results probably because end position might have been smaller than the start position and this error results in making a mistake on the whole answer, DCA model gets confused as ``central business districts"  term also doesn't contain downtown.\\
    \hline
    \bfseries Context 			  &................ , the most important of which were use of a nitrogen/oxygen mixture instead of pure oxygen before and during launch, and removal of flammable cabin and space suit materials. the block ii .........\\ 
    \bfseries Question    		  & what type of materials inside the cabin were removed to help prevent more fire hazards in the future?\\
    \bfseries True Answer   	  	  & flammable cabin and space suit materials\\
    \bfseries Predicted Answer   	  &  space suit materials (Hybrid model) / flammable cabin and space suit materials(DCA model)\\
    \bfseries Explanation of Model Behavior   & Hybrid model was not able to capture the ``flammable cabin" part probably because it was followed by cabin and it looked for materials following the word ``cabin". DCA captures this correctly. Technically both answers are correct\\
    
    \hline
    \bfseries Context 			  & .............., the statistical behaviour of primes in the large, can be modelled. the first result in that direction is the prime number theorem, proven at the end of the 19th century, which says that.........\\ 
    \bfseries Question    		  & what theorem states that the probability that a number n is prime is inversely proportional to its logarithm ?\\
    \bfseries True Answer   	  	  & the prime number theorem\\
    \bfseries Predicted Answer   	  & the prime number theorem (Hybrid model) / prime number theorem , proven at the end of the 19th century(DCA model)\\
    \bfseries Explanation of Model Behavior   & While Hybrid model gets it right, DCA answer is not wrong but a model trained on tighter start and end dependencies might have been able to capture this more accurately.\\
    \hline
    \end{tabular}
}
\hfill{}
\label{tb:error}
\end{table}
\end{center}

\section{Conclusions and Future work}
In this paper, we studied and implemented two well known attention mechanisms namely, BiDAF and Co-Attention. We also introduced a simple combination of these two schemes called Hybrid attention mechanism that outperforms both BiDAF and Co-Attention. In addition to this, we propose our own attention mechanism called Double Cross Attention that gives similar results as the Hybrid model. The objective of the paper was primarily to study and compare the two aforementioned popular attention schemes on their own and not to chase the leaderboard scores. In particular, we isolated the attention layer and suggested our own improvements to it. The comparative results between different schemes are obtained for same set of hyperparameters. 

To improve the F1/EM scores of the overall system, a number of enhancement techniques could be used. For e.g. while we simply concatenated character and word embeddings, more advanced techniques to effectively combine them have been suggested in the literature \cite{gating}. Also. a number of other attention mechanisms have been suggested which need to be investigated as well \cite{match, aoa}. Another possible improvement is to properly condition the end position on the start position of the answer span. An LSTM based solution was used in the original BiDAF paper. Exponential moving average of weights and ensembling are additional common methods to further fine-tune and improve the results. Hierarchical Maxout Network as mentioned in the co-attention paper could be a replacement to our simple Softmax output layer to improve the performance even further. There are also a few possible directions where DCA model can further be improved/extended. We can continue recursively calculating the cross attention weights and combine them in some more intuitive or non-linear way. While, we didn't optimize for the number of parameters, it is possible to reduce the overall number of trainable parameters by appropriately sharing weights between layers when possible.

All of the above mentioned suggestions, we see as enhancement opportunities (some we partially already tried to implement but could not finally manage to include in final running model). As a final project for the cs224n course, we found the task challenging but we were extremely satisfied with our own personal learning curve. We are sure that with even more time, we could significantly improve our model from the baseline enhancement we achieved so far. All in all, we believe that the experience of this project, will be of utmost value for our future professional work. 

%
%
%
%

\section*{Acknowledgements}
First of all, we would like to thank the course instructor Richard Socher for making the class highly informative and a great learning experience. We would also like to thank the TAs for prompt feedback and insightful discussions. Lastly, we would like to thank the fellow students who regularly helped each other on course forum regarding any questions..

\end{document}